\documentclass[conference]{IEEEtran}
\IEEEoverridecommandlockouts
\usepackage{cite}
\usepackage{algpseudocode,algorithm,algorithmicx}
\usepackage{amsmath,amssymb,amsfonts}
\usepackage{amsthm}
\usepackage{graphicx}
\usepackage{booktabs}
\usepackage{textcomp}
\usepackage{xspace}
\usepackage{subcaption}
\usepackage{caption}
\usepackage{color}
\usepackage{multirow}
\usepackage{hyperref}
\newcommand{\eg}{\textit{e}.\textit{g}.\@\xspace}
\newcommand{\ie}{\textit{i}.\textit{e}.\@\xspace}
\newcommand{\etal}{\textit{et al}.\@\xspace}

\newtheorem{hypothesis}{Hypothesis}
\newtheorem{lemma}{Lemma}

\def\mbx{x}
\def\mbz{z}
\def\mcf{\mathcal{F}}
\def\mck{\mathcal{K}}

\begin{document}

\title{FedZKT: Zero-Shot Knowledge Transfer towards Resource-Constrained Federated Learning with Heterogeneous On-Device Models\\

}

\author{\IEEEauthorblockN{Lan Zhang}
\IEEEauthorblockA{\textit{Department of Electrical and Computer} \\
\textit{Engineering, Michigan Technological University}\\
Houghton, MI USA \\
lanzhang@mtu.edu}
\and
\IEEEauthorblockN{Dapeng Wu}
\IEEEauthorblockA{\textit{Department of Electrical and Computer} \\
\textit{Engineering, University of Florida}\\
Gainesville, FL USA \\
dpwu@ufl.edu}
\and
\IEEEauthorblockN{Xiaoyong Yuan}
\IEEEauthorblockA{\textit{College of Computing} \\
\textit{Michigan Technological University}\\
Houghton, MI USA \\
xyyuan@mtu.edu}
}

\maketitle

\begin{abstract}
Federated learning enables multiple distributed devices to collaboratively learn a shared prediction model without centralizing their on-device data. Most of the current algorithms require comparable individual efforts for local training with the same structure and size of on-device models, which, however, impedes participation from resource-constrained devices. Given the widespread yet heterogeneous devices nowadays, in this paper, we propose an innovative federated learning framework with heterogeneous on-device models through Zero-shot Knowledge Transfer, named by FedZKT. Specifically, FedZKT allows devices to independently determine the on-device models upon their local resources. To achieve knowledge transfer across these heterogeneous on-device models, a zero-shot distillation approach is designed without any prerequisites for private on-device data, which is contrary to certain prior research based on a public dataset or a pre-trained data generator. Moreover, this compute-intensive distillation task is assigned to the server to allow the participation of resource-constrained devices, where a generator is adversarially learned with the ensemble of collected on-device models. The distilled central knowledge is then sent back in the form of the corresponding on-device model parameters, which can be easily absorbed on the device side. Extensive experimental studies demonstrate the effectiveness and robustness of FedZKT towards on-device knowledge agnostic, on-device model heterogeneity, and  other challenging federated learning scenarios, such as heterogeneous on-device data and straggler effects.
\end{abstract}

\vspace{1em}

\section{Introduction}
The demand for on-device training is recently increasing, as evinced by the surge of interest in federated learning~\cite{yang2019federated}. Federated learning leverages on-device training at multiple distributed devices to obtain a knowledge-abundant global model without centralizing private on-device data~\cite{yang2019federated,mcmahan2017communication}. Classical federated learning algorithms, represented by FedAvg~\cite{mcmahan2017communication}, require on-device training with the same model structure and size to perform the element-wise central average, which, however, impedes collaboration across {heterogeneous hardware platforms}. For instance, both wearable devices and smartphones are popular eHealth devices to monitor infectious diseases~\cite{chen2021age}. However, a typical wearable device is usually equipped with microcontroller units (MCU), whose on-chip memory is three orders of magnitude smaller than a smartphone due to the limited resource budget, especially the memory (SRAM) and storage (Flash)~\cite{lin2020mcunet}. Hence, a wearable device can hardly run the same on-device model designed for a smartphone~\cite{cai2019once}, resulting in the ineffectiveness of implementing classical federated learning. 
Given the widespread yet heterogeneous devices nowadays, it is vital to enable extensive participation in federated learning, especially with resource-constrained devices.

One promising solution is to allow federated learning with heterogeneous on-device models, which recently has attracted great attention. Diao \etal proposed HeteroFL to adaptively allocate a subset of global model parameters to an on-device model~\cite{diao2020heterofl}. Inherently, HeteroFL assumes the architecture of a small model can be a subnetwork of a large one, which, however, is not always practical. For example, it is hard to find the architecture of MobileNet~\cite{sandler2018mobilenetv2}, \ie, a popular on-device model, as a subnetwork of other models, such as ShuffleNet~\cite{ma2018shufflenet} and ResNet~\cite{he2016deep}. Instead, some recent research enabled devices to design their on-device models independently based on federated distillation techniques~\cite{li2019fedmd,chang2019cronus,li2021fedh2l,lin2020ensemble}. Specifically, the logit information of on-device models is shared in FedMD~\cite{li2019fedmd}, Cronus~\cite{chang2019cronus}, and FedH2L~\cite{li2021fedh2l} to achieve federated learning for personalization, security, and decentralization, respectively, and on-device model parameters are shared in FedDF~\cite{lin2020ensemble} for robust model fusion. Although successful, all above algorithms rely on certain prerequisites of on-device knowledge, which leverage either a public dataset or a pre-trained data generator to extract and transfer knowledge. Consequently, the construction of such data-dependency requires careful deliberation and even prior knowledge of the private on-device data, making it infeasible in many applications in practice. Moreover, the impact of the quality of such prerequisites remains unclear on federated learning performance.  

To tackle the above limitations of existing research, in this paper, we propose a \underline{Z}ero-shot \underline{K}nowledge \underline{T}ransfer framework named by FedZKT for resource-constrained federated learning with heterogeneous on-device models in a data-free manner. Specifically, in FedZKT, devices can design on-device models based on their heterogeneous local resources independently. To enable knowledge transfer across these on-device models, a zero-shot federated distillation approach is proposed without any prerequisite for private on-device data, which is contrary to the aforementioned prior research. This compute-intensive distillation task is assigned to the server to reduce the workload at devices, where the server constructs a generator to be adversarially trained with the ensemble of the collected on-device models. The distilled central knowledge is then sent back in the form of the corresponding on-device model parameters, which can be easily absorbed on the device side. In other words, FedZKT enables extensive participation, especially from the ubiquitous resource-constrained devices, who can easily contribute to federated learning by following the classical federated learning procedures with locally designed compact models. Overall, FedZKT consolidates several \textit{advantages} into a single framework: independent on-device model design, extensive participation from resource-constrained and/or heterogeneous devices, and data-free knowledge transfer. Our main contributions are as follows: 
\begin{itemize}
\item This paper introduces an innovative framework, FedZKT, for resource-constrained federated learning in a data-free manner, which performs zero-shot knowledge transfer across heterogeneous on-device models. Several key modules are designed to implement the lightweight and compute-intensive learning tasks on the device and server sides, respectively, which perfectly fits the unbalanced resources on both sides.  
\item Contrary to certain prior research based on either a public dataset or a pre-trained data generator, FedZKT provides an on-device knowledge agnostic approach without data-dependency concerns, where the server adversarially learns a generative model with the global model based on the ensemble of collected on-device models. A new loss function, softmax $l_1$ (SL) loss, is proposed to facilitate the zero-shot federated knowledge distillation. 
\item Extensive experimental results demonstrate the effectiveness of FedZKT on four popular datasets, with higher accuracy and better generalization performance compared to the state-of-the-art. FedZKT also performs robustness to challenging federated learning scenarios, such as non-iid data distribution and straggler effects. 
\end{itemize}

\section{Related Work}
\subsection{Heterogeneous Federated Learning.}\label{sec:}
Classical federated learning algorithms, represented by FedAvg~\cite{mcmahan2017communication}, average the collected on-device model parameters to obtain a global model. Since the training mainly happens on the device side, the overall learning performance largely depends on participating devices. It has been shown that the statistical heterogeneity across devices, \ie, non-iid on-device data, can lead to slow and unstable convergence~\cite{li2019convergence,zhao2018federated}. Such performance degradation has also been found when on-device resources, such as the local computing power or network connectivity, are heterogeneous~\cite{li2020federated}.  Recent research has developed solutions to either address the ``straggler effect'' introduced by some poorly performed devices~\cite{liu2020accelerating,mcmahan2017communication,li2020federated,nishio2019client} or reduce the local model size at all devices~\cite{vepakomma2018split,he2020group}. However, most of these designs are still under the learning paradigm of FedAvg with homogeneous on-device models, \ie, all devices need to run on-device models with the same structure and size. 

\vspace{-0.2em}
\subsection{Federated Distillation.}\label{sec:RW_FD}
To allow federated learning with heterogeneous on-device models, federated distillation has attracted significant attention recently. Enlighten by the well-known knowledge distillation idea that transfers knowledge from a single or multiple teacher models to an empty student model~\cite{bucilu2006model,ba2013deep,hinton2015distilling,cho2019efficacy}, federated distillation learns a global model based on the collected on-device models. Since the data is stored locally and cannot be shared in federated settings, most federated distillation design is data-dependent. Specifically, by leveraging a pre-known public or surrogate dataset, federated distillation has been studied to handle heterogeneous on-device models for personalization~\cite{li2019fedmd}, security~\cite{chang2019cronus}, decentralization~\cite{li2021fedh2l}, and robustness~\cite{lin2020ensemble}, respectively. In addition, federated distillation has been used to improve federated learning performance, such as communication efficiency~\cite{jeong2018communication,guha2019one,itahara2020distillation,sattler2020communication,seo2020federated} and on-device privacy~\cite{sun2020federated,chang2019cronus}, or address the aforementioned data heterogeneity challenges~\cite{chen2020feddistill,sattler2021fedaux}. However, the prerequisite of the data-dependency is not always available for federated distillation due to the unknown or confidential data distribution of on-device models. Moreover, the absence of a qualified public or surrogate dataset can return a poor approximation of teacher models in knowledge distillation~\cite{truong2021data}, whose impact on federated learning still remains unclear. In this paper, we target a data-free federated distillation design without the data-dependency prerequisite for the resource-constrained federated learning with heterogeneous on-device models.


\subsection{Data-Free Knowledge Distillation.}\label{sec:RW_DFKD}
Recent efforts have been made on data-free knowledge distillation via zero-shot learning techniques~\cite{fang2019data,micaelli2019zero,choi2020data,truong2021data}. Typically, a generative model is learned to synthesize the queries that the student makes to the teacher. Although this idea has been well studied in classical knowledge distillation, such as for model compression~\cite{fang2019data}, little attention has been paid to distillation in federated settings. To the best of our knowledge, the data-free distillation design in FeDGen~\cite{zhu2021data} is the closest setting to this work. Specifically, FeDGen targets the slow convergence issue due to data heterogeneity in federated learning, while this paper aims to enable resource-constrained federated learning with heterogeneous on-device models. In FeDGen, a generator is used on the device side to augment local knowledge for data-free distillation. Instead of distributing the generator to devices, this paper learns and keeps the generator at the server. Only the updated on-device model parameters will be sent back to devices, which can be easily absorbed locally. In this way, the compute-intensive data-free distillation task is assigned to the powerful server to enable extensive participation especially from resource-constrained devices.


\begin{figure}[!tb]
\centering
\includegraphics[width=\linewidth]{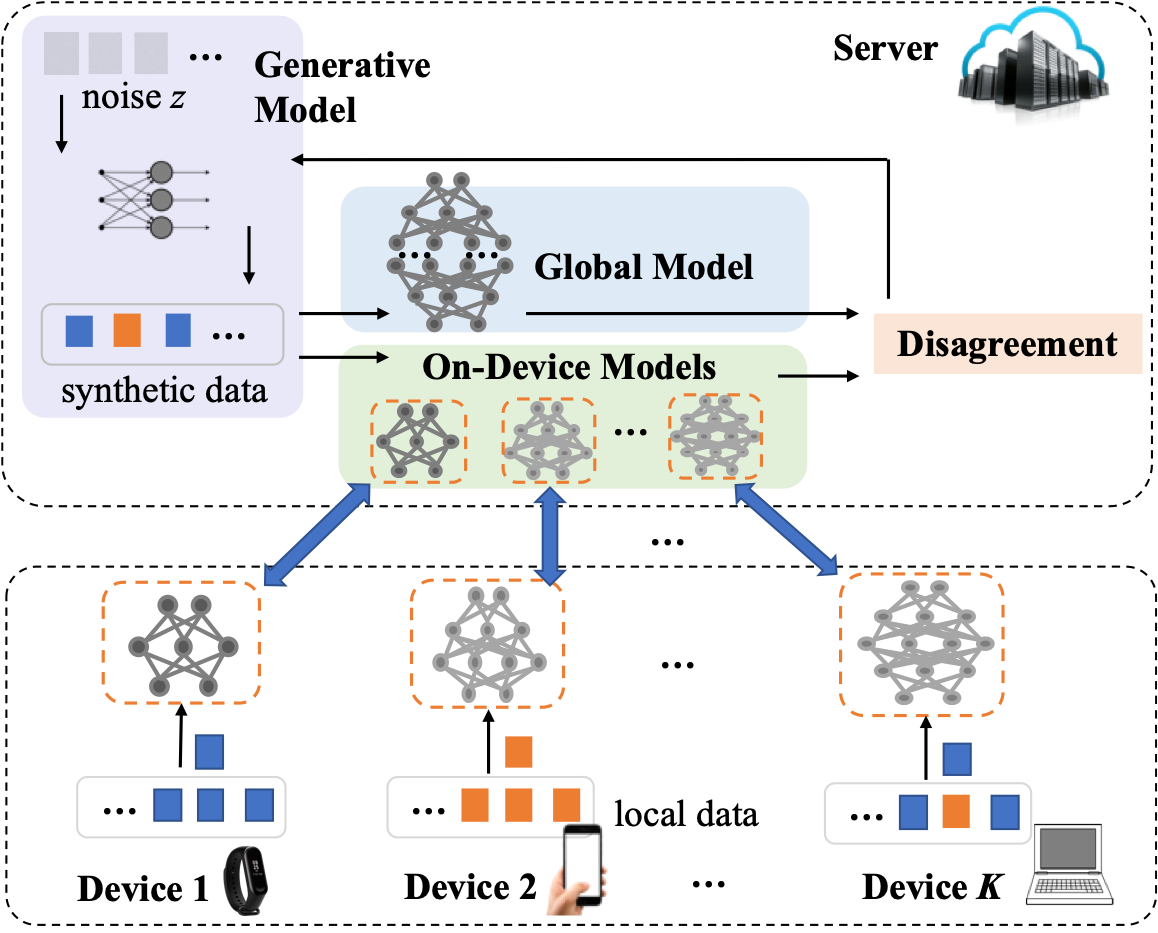}
\caption{\textbf{Overview of FedZKT: zero-shot knowledge transfer for federated learning with heterogeneous on-device models.} Most compute-intensive workloads of FedZKT are done at the server. Participating devices only need to update the on-device models based on their local data.} 
\label{fig:system}
\end{figure}

\section{FedZKT: Federated Learning via Zero-shot Knowledge Transfer}
This section presents the proposed FedZKT. We first describe the problem statement, followed by FedZKT design. Several key modules of FedZKT are detailed at both the server and device sides.

\subsection{Problem Statement}
As shown in Figure~\ref{fig:system}, we consider a federated learning task, \eg, a supervised classification task, across $K$ heterogeneous devices in set $\mck$. Each device $k\in\mck$ can independently design its on-device model $f_k$ parameterized by $w_k$ to ``best'' fit its own resources in computation, communication, storage, and power. In this case, the on-device models $\{f_k\}_{k\in\mck}$ may have distinct model architectures. In addition, we assume each device owns a confidential dataset $\mathcal{D}_k$ that cannot be shared, and the server has no prior knowledge about the data at the device side, such as the data distribution. The server is assumed to be powerful, who coordinates all participating devices and aggregates their on-device knowledge. Hence, the main goal of the server is to extract distributed on-device knowledge from heterogeneous on-device models $\{f_k\}_{k\in\mck}$ to obtain a knowledge-abundant global model $F$. Meanwhile, the well-trained on-device models $\{f_k\}_{k\in\mck}$ will eventually contain the distilled knowledge from all peer devices. 

\subsection{FedZKT Design}
To meet the unbalanced computing capabilities between the server and participating devices, the compute-intensive knowledge distillation task is assigned to the server. The procedures of the overall FedZKT are introduced in Algorithm~\ref{alg:fedzkt}. In the following, we will elaborate design principles of several key modules of FedZKT.

\begin{algorithm}[!tb]
\caption{FedZKT }\label{alg:fedzkt}
\textbf{INPUT:} global model parameters $w$, on-device model parameters $\{w_k\}_{k\in\mck}$, generative model parameters $\theta$, total communication rounds $T$, local training epochs $T_l$, distillation epochs $n_D$.
\begin{algorithmic}[1]
\State Initialize all models with random weights\footnotemark.
\For{each communication round $t=1,2,\cdots,T$}
\State $\mck^t \leftarrow$ server selects a random subset devices from $\mck$ as active devices
\vspace{0.5em}
\State // \textit{On-Device Update}
\For{each device $k\in\mck^t$ \textbf{in parallel}}
    \State $\hat{w}_k \leftarrow$ \textbf{DeviceUpdate}($w_k$, $\mathcal{D}_k$, $T_l$, $n_D$) 
    \State Upload $\hat{w}_k$ to the server.
\EndFor
 
\vspace{0.5em}
\State // \textit{Server Update}
\State $\{w_k\}_{k\in\mck}, w, \theta$ $\gets$ \textbf{ServerUpdate}($\{\hat{w}_k\}_{k\in\mck},w,\theta)$.
\For{each device $k\in\mck$ \textbf{in parallel}}
    \State Transfer $w_k$ to Device $k$.
\EndFor
\EndFor
\end{algorithmic}
\textbf{RETURN:} $w$, $\{w_k\}_{k\in\mck}$
\end{algorithm}
\footnotetext{This work uses Glorot initialization~\cite{glorot2010understanding}. The same initialization is not required for on-device models.}

\begin{algorithm}[!tb]
\caption{FedZKT: \textbf{DeviceUpdate}}\label{alg:client}
\textbf{INPUT:} on-device model $f_k$'s parameters $w_k$, local dataset $\mathcal{D}_k$, local epochs $T_l$
\begin{algorithmic}[1]
\For{$t \in \{1,2,...,T_l\}$}
    \State $\mathcal{L}_k \gets \sum_{\{\mbx, y\} \in \mathcal{D}_k}\mathcal{L}_{CE}(f_k(\mbx;w_k), y)$ 
    \State $w_k \gets w_k - \eta \frac{\partial \mathcal{L}_k}{\partial w_k}$
\EndFor
\end{algorithmic}
\textbf{RETURN: $w_k$}
\end{algorithm}

\begin{algorithm}[!tb]
\caption{FedZKT: \textbf{ServerUpdate}}\label{alg:server}
\textbf{INPUT:} global model $\mcf$'s parameters $w$,  on-device model $f_k$'s parameters $\{w_k\}_{k\in\mck}$, generator $G$'s parameters $\theta$, distillation epochs $n_D$.
\begin{algorithmic}[1]
\State // \textit{Transfer knowledge from on-device to the global model}\vspace{0.1em}
\For{$n \in \{1,2,...,n_D\}$}
    \State // \textit{Generator Update}
    \State $\mbz \sim \mathcal{N}(0,\mathbf{I})$
    \State $\mbx \gets G(\mbz; \theta)$
    \State $\mathcal{L}_G \gets - \mathcal{L}(\mcf(\mbx; w), \frac{1}{|\mck|}\sum_{k\in\mck}f_k(\mbx; w_k))$
    \State $\theta \gets \theta - \eta_G \frac{\partial \mathcal{L}_G}{\partial \theta}$
    \vspace{0.5em}
    \State // \textit{Global Model Update}\vspace{0.1em}
    \State $\mbz \sim \mathcal{N}(0,\mathbf{I})$
    \State $\mbx \gets G(\mbz; \theta)$
    \State $\mathcal{L}_S \gets \mathcal{L}(\mcf(\mbx; w), \frac{1}{|\mck|}\sum_{k\in\mck}f_k(\mbx; w_k))$
    \State $w \gets w - \eta_S \frac{\partial \mathcal{L}_S}{\partial w}$
\EndFor
\vspace{0.5em}
\State // \textit{Transfer knowledge from global to on-device models}
\For{$n \in \{1,2,...,n_D\}$}
    \State $\mbz \sim \mathcal{N}(0,\mathbf{I})$
    \State $\mbx \gets G(\mbz; \theta)$
    \For{each device $k\in\mck$}
        \State ${\mathcal{L}_k} \gets \mathcal{L}(\mcf(\mbx; w), f_k(\mbx; w_k))$ 
        \State $w_k \gets w_k - \eta_S \frac{\partial \mathcal{L}_k}{\partial \hat{w}_k}$
    \EndFor
\EndFor
\end{algorithmic}
\textbf{RETURN:} $\{w_k\}_{k\in\mck}$, $w$, $\theta$
\end{algorithm}

\vspace{0.3em}
\subsubsection{Zero-Shot Knowledge Distillation}\label{sec:zero-shotDist}
As aforementioned, the goal of knowledge distillation at the server is to obtain the global model $\mcf(\cdot; w)$ parameterized by $\omega$, which is expected to match the ensemble of on-device models $\{f_k\}_{k\in\mck}$ that are trained respectively on distributed local knowledge domains $\{\mathcal{D}_k\}_{k\in\mck}$. Since $\{\mathcal{D}_k\}_{k\in\mck}$ are private and cannot be shared in federated settings, one intuitive idea is to leverage a synthetic dataset $\mathcal{D}_\mathcal{S}$ to mimic the local knowledge in order to minimize the loss of disagreement between the teacher and the students. Thus, based on ensemble learning, we have
\begin{equation}
\label{equ_KDLoss}
\min_w E_{\mbx\sim\mathcal{D}_\mathcal{S}} [\mathcal{L}(\mcf(\mbx; w), f_{\mathrm{ens}}(x))],
\end{equation}
where $f_{\mathrm{ens}}$ denotes the ensemble of on-device models, \ie, $f_{\mathrm{ens}}(x) =\frac{1}{|\mck|}\sum_k f_k(\mbx; w_k)$, and $\mathcal{L}$ denotes the loss function to measure the disagreement between the global model $\mcf$ and $f_{\rm ens}$. More discussions about $\mathcal{L}$  will be given in the next module.  

Since $\mathcal{D}_\mathcal{S}$ is expected to synthesize the private local knowledge in a data-free manner, instead of assuming a pre-known $\mathcal{D}_S$, we introduce a generative model $G$ to distill local knowledge in a zero-shot manner. Enlighten by the well-known idea of Generative Adversarial Networks (GAN)~\cite{truong2021data}, $G$ is responsible to provide difficult inputs for the training of $\mcf$, which maximizes the disagreement between the current global and on-device models. Meanwhile the generated inputs also need to perform well in (\ref{equ_KDLoss}) to enable knowledge matching between the global and on-device models. Hence, the goals of $G$ and $\mcf$ are to maximize and minimize the disagreement between $\mcf$ and $\{f_k\}_{k\in\mck}$, respectively, where the adversarial game can be given by 
\begin{equation}\label{equ_minmaxKD}
\min_{\mcf}\max_{G} E_{\mbz\sim\mathcal{N}(0,1)}[\mathcal{L}(\mcf(G(\mbz)), f_{\mathrm{ens}}(G(\mbz)))],
\end{equation}
where $\mbz$ is the noise following Gaussian distribution $\mathcal{N}(0,1)$. Therefore, the server will alternatively train the generative model $G$  and global model $\mcf$ in (\ref{equ_minmaxKD}). 
It should be mentioned that most prior knowledge distillation approaches perform well when extracting a small student model from a large teacher model, while some recent research~\cite{micaelli2019zero,truong2021data} has shown that high distillation accuracy can be achieved even when the teacher model has a smaller and different architecture than the student's. Along this line, this work is further motivated to transfer knowledge between the powerful global model and the heterogeneous and potentially compact on-device models. 

\vspace{0.3em}
\subsubsection{Loss Function Design}\label{sec:loss_kl}
This module discusses the loss function $\mathcal{L}$ in (\ref{equ_minmaxKD})  that measures the disagreement between the global model $\mcf$ and the on-device model ensemble $f_{\rm ens}$, which is used to train $\mcf$ and the generative model $G$ simultaneously. The loss function design is key to the distillation performance since the gradients computed through $\mcf$ and  $f_{\rm ens}$ can easily impede the convergence of the optimizer, such as leading to gradient vanishing~\cite{fang2019data} when the wrong loss function is used.

Most prior research of knowledge distillation measures the model disagreement between the teacher and the student by \textit{Kullback–Leibler (KL) divergence}~\cite{cho2019efficacy,hinton2015distilling}. Hence, the KL divergence between the outputs of the global model $\mcf$ and the ensemble of on-device models $f_{\rm ens}$ after the softmax function becomes a candidate for the loss function, where the KL-divergence loss function can be given by 
\begin{equation}
    \mathcal{L}_{\mathrm{KL}}(\mbx) = \sum\mcf(\mbx) \log \frac{\mcf(\mbx)}{f_{\mathrm{ens}}(\mbx)}.
\end{equation}
However, the KL-divergence loss tends to suffer from gradient vanishing~\cite{fang2019data} with respect to input data $\mbx$ when the student model $\mcf$ converges to the teacher model $f_{\rm ens}$. The problem becomes even more serious in zero-shot distillation settings, since the gradient vanishing will further affect the training of the generative model $G$. 

In view of this, recent zero-shot distillation research~\cite{fang2019data,truong2021data} introduces $\ell_1$ \textit{norm loss}, which compares the logit outputs (model outputs before the softmax layer) between the teacher and student models:
\begin{equation}
    \mathcal{L}_{\ell_1}(\mbx) = ||u(\mbx)-\frac{1}{|\mck|}\sum_k v_k(\mbx)||_1,
\end{equation}
where $u$ and $v_k$ denote the logit outputs of the global model (student) and the $k$th on-device model (teacher), respectively. However, given the diverse on-device model parameters in our heterogeneous federated learning, the $\ell_1$ norm loss may lead to the unstable training due to the large gradients. Specifically, federated learning requires aggregating distributed knowledge from participating devices. However, averaging the logit values over on-device models increases the gradients, making the whole learning process unstable. 

To address the above challenges for zero-shot distillation in federated learning, we propose a new loss function named by \textit{Softmax} $\ell_1$ \textit{(SL) loss}, which applies the softmax output to the $\ell_1$ norm loss:
\begin{equation}
    \mathcal{L}_{\mathrm{SL}}(\mbx) = ||\mcf(\mbx) - f_{\mathrm{ens}}(\mbx)||_1.
\end{equation}
The SL loss is designed to overcome the drawbacks of using KL-divergence loss and $\ell_1$ norm loss. Two hypotheses for this design are provided below. Specifically, Hypothesis~\ref{hp:kl} suggests that the SL loss can reduce the gradient vanishing effect than the KL-divergence loss for better convergence in zero-shot distillation; Hypothesis~\ref{hp:l1} suggests that the SL loss can make the training more stable compared to the $\ell_1$ norm loss. 
Details are given in the Appendix for justification. In addition, Figure~\ref{fig:gradient} shows the norm of gradients for KL-divergence, $\ell_1$ norm, and the proposed SL norm losses, respectively, where the gradients for the KL-divergence loss tend to vanish, while the gradients for the $\ell_1$ norm loss are much larger and unstable during the learning process. More empirical evaluation results are given in Section~\ref{sec:loss}.  

\begin{figure}[!tb]
\centering
\includegraphics[width=0.7\linewidth]{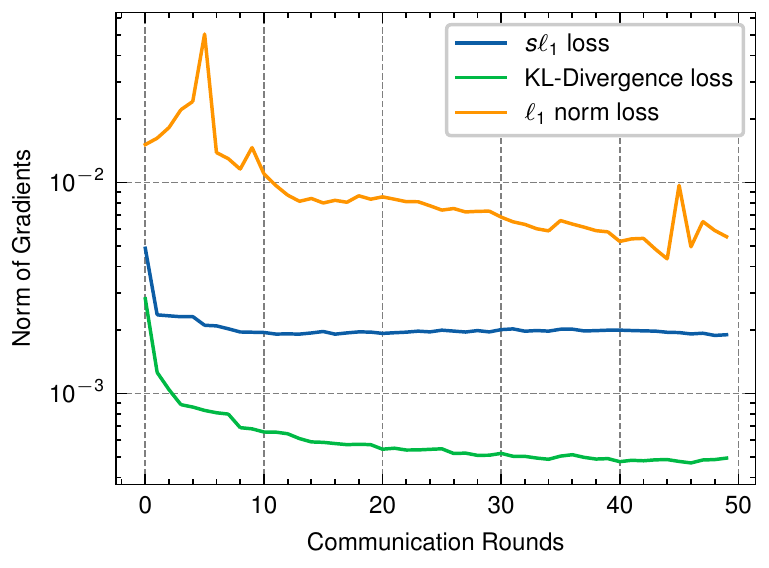}
\caption{\textbf{Norm of gradients w.r.t input data (MNIST, IID).} The gradients for the KL-divergence loss tend to vanish, while the gradients for the $\ell_1$ norm loss are much larger and unstable during the learning process. The proposed SL loss overcomes both problems in the federated learning.}
\label{fig:gradient}
\end{figure}

\begin{hypothesis}\label{hp:kl}
When the global model $F$ converges to the ensemble of on-device models $f_{\mathrm{ens}}$, the gradients of KL divergence loss with respect to the input data $\mbx$ are smaller than those of the SL loss:
\begin{equation}
||\nabla_{\mbx}\mathcal{L}_{\mathrm{KL}}(\mbx)|| \underset{F\rightarrow f_{\mathrm{ens}}}{\leq}  ||\nabla_{\mbx}\mathcal{L}_{\mathrm{SL}}(\mbx)||.
\end{equation}
\end{hypothesis}

\begin{hypothesis}\label{hp:l1}
When the global model $F$ converges to the ensemble of on-device models $f_{\mathrm{ens}}$, the gradients of the $\ell_1$ norm loss with respect to the input data $\mbx$ are greater than those of the SL loss:
\begin{equation}
||\nabla_\mbx\mathcal{L}_{\ell_1}(\mbx)|| \underset{F\rightarrow f_{\mathrm{ens}}}{\geq}  ||\nabla_{\mbx}\mathcal{L}_{\mathrm{SL}}(\mbx)||.
\end{equation}
\end{hypothesis}

\vspace{0.3em}
\subsubsection{Bidirectional Knowledge Transfer}
The above two modules enable the knowledge transfer from devices to the server. After that, the aggregated central knowledge needs to transfer back for the next learning iteration. One intuitive approach is to broadcast the updated global model $\mcf$, based on which, device $k$ can use its data $\mathcal{D}_k$ to distill an updated on-device model, $\min_{w_k} E_{x\sim\mathcal{D}_k} [\mathcal{L}(\mcf(x), f_k(x; w_k))]$, similar to our discussion in the above module based on (\ref{equ_KDLoss}). However, since our design aims to enable resource-constrained federated learning, to utmostly reduce the workload at devices, we run the compute-intensive distillation task to transfer knowledge from the global model to on-device models at the server. Since the above module learns the generator $G$ to produce difficult data that maximizes the disagreement between global and on-device models, we will reuse this well-learned generator $G$ to provide input data to distill the updated on-device model $\{f_k'\}_{k\in\mck}$. Hence, the objective function for knowledge transfer from the global model $\mcf$ to the on-device model $f_k'$, $k\in\mathcal{K}$ can be given by
\begin{equation}\label{equ_biKD}
\min_{f_k'} E_{\mbz\sim\mathcal{N}(0,1)} [\mathcal{L}(\mcf(G(\mbz)), f_k'(G(\mbz)))].
\end{equation}
Different from the distillation of module~\ref{sec:zero-shotDist} designed for a zero-shot setting, we adopt the KL-divergence loss $\mathcal{L}_{\rm KL}$ here for distillation with a pre-trained data generator $G$. The updated on-device model $f_k'$ will be sent back to device $k$, which obtains the central knowledge learned by global model $\mcf$. This completes the one-round bidirectional knowledge transfer. 
Since the computation of (\ref{equ_minmaxKD}) and (\ref{equ_biKD}) is performed at the server, FedZKT follows the same on-device learning mode as classical federated learning~\cite{mcmahan2017communication}, while allowing participation with independently designed compact on-device models.

\vspace{0.5em}
\subsubsection{$\ell_2$ Regularization for Non-IID Data Distribution}\label{sec:l2reg}
In addition to tackling the model architecture heterogeneity, FedZKT is expected to handle data heterogeneity in federated learning. To do so, we limit the update of on-device models when training on their local datasets in Algorithm~\ref{alg:client}. Specifically, the $\ell_2$ regularization is added to the loss function of the on-device update, which is given by
\begin{equation}
    \min_{w_k^t} \sum_{\{\mbx, y\} \in \mathcal{D}_k}\mathcal{L}_{CE}(f_k(\mbx;w_k^t), y) + ||w_k^t - w_k^{t-1}||_2^2,
\end{equation}
where $\mathcal{L}_{CE}$ is the cross-entropy loss for classification tasks, and $w_k^{t-1}$ is the parameter set transferred from the server in the last iteration $t-1$. 
The $\ell_2$ regularization has been used to tackle the non-iid data distribution with homogeneous model architectures in FedProx~\cite{li2018federated}. 
Compared with FedProx, due to model heterogeneity,  FedZKT uses the received local model parameter set $w_k^{t-1}$ rather than the global model parameter set $w^{t-1}$. Empirical evaluation in Section \ref{sec:l2_reg} further demonstrates the improvement of adding $\ell_2$ regularization for non-iid data distributions.

\section{Experimental Validation}
This section conducts extensive experiments to evaluate the proposed FedZKT.  We first introduce the experimental setup, followed by the experimental results and ablation studies. 

\subsection{Experimental Setup}
\subsubsection{Dataset} The experiments are conducted on four widely used image datasets: MNIST~\cite{lecun1998gradient}, KMNIST~\cite{clanuwat2018deep}, FASHION-MNIST~\cite{xiao2017fashion} (FASHION in short in this paper), and CIFAR-10~\cite{krizhevsky2009learning}.

\subsubsection{On-Device Model Heterogeneity} 
Five different neural network architectures are considered for each dataset. For small datasets, \ie, MNIST, KMNIST, and FASHION, we deploy a CNN model, a Fully-Connected Model, and three LeNet-like models with different channel sizes and numbers of layers. For CIFAR-10 dataset, we deploy two ShuffleNetV2 models~\cite{ma2018shufflenet}, two MobileNetV2 models~\cite{sandler2018mobilenetv2}, and a LeNet-like model. Specifically, to support resource-constrained federated learning, ShuffleNetV2 and MobileNetV2 are adopted, which are popular neural architectures designed towards low-end devices. LeNet is a simple neural network architecture consisting of two convolutional layers and three fully connected layers. We use different channel sizes for each ShuffleNetV2 and MobileNetV2 model, so as to increase the heterogeneity of on-device models in the evaluation.

\subsubsection{Federated Learning Settings} The experiments are conducted with multiple devices $K\in\{5, 10, 15, 20\}$ (by default $k$=10). For the small datasets, \ie, MNIST, KMNIST, FASHION, we conduct 50 communication rounds ($T=50$); in each round, each device trains the on-device models for 5 epochs. For the CIFAR-10 dataset, we conduct 100 communication rounds ($T=100$); in each round, each device trains the on-device model for 10 epochs. We use the stochastic gradient descent (SGD) for both the on-device and global training, where the learning rate is set to be 0.01. 
Besides, we train the server model and the generator for 200 iterations for the small datasets ($n_G=n_s=200$) and 500 iterations for the CIFAR-10 dataset ($n_G=n_s=500$). The generator is trained using Adam optimizer with a batch size of 256 and a learning rate of 0.001. The learning rates for both the server model and the generator are reduced by 0.3 at the half and 3/4 of the total iterations. The batch size for all model training is 256.

\subsubsection{Data Heterogeneity} \label{sec:exp:datahetero}
The experiments will be conducted on both iid and non-iid on-device data distributions. In the iid setting, on-device data is randomly drawn from the dataset. In the non-iid setting, two scenarios of label distribution skew are adopted based on the common experimental settings in recent federated learning research~\cite{wang2020tackling,li2021federated}: 1) quantity-based label imbalance, where each device owns data consisting of a specific number of classes;
2) distribution-based label imbalance, where each device owns a proportion of the labels following a Dirichlet distribution. We sample $p_k = \{p_{kj}\}$ from $Dir_N(\beta)$, where $p_{kj}$ denotes the proportion of data in class $k$ owned by device $j$, and $\beta$ is a concentration parameter of the Dirichlet distribution. A small value of $\beta$ suggests a more imbalanced distribution of labels.  

\subsubsection{Baseline Approach}
As aforementioned in Section \ref{sec:RW_FD} and \ref{sec:RW_DFKD}, existing federated learning designs to support heterogeneous on-device models are data-dependent, \ie, based on either a public dataset or a pre-trained generator, while the proposed FedZKT is based on a data-free manner. Hence, the baseline approach for FedZKT adopts the most representative data-dependent algorithm, FedMD~\cite{li2019fedmd}. Similar to FedZKT, FedMD allows devices to independently design their on-device models, which provides extensive experimental results for on-device model heterogeneity but requiring a public dataset. To learn from MNIST,  FASHION, and KMNIST, we select public datasets as FASHION, MNIST, and FASHION, respectively. To explore the impact of data dependency during knowledge transfer, we select two different public datasets, \ie, CIFAR-100 and SVHN, to learn CIFAR-10. 

\subsection{Experimental Results}
\subsubsection{IID Data Distribution}

\begin{table}[!tb]
\centering
\begin{tabular}{@{}lllr@{}}
\toprule
On-Device Dataset & \multicolumn{2}{c}{FedMD} & \multicolumn{1}{c}{FedZKT} \\ \midrule
 & \multicolumn{1}{l}{\begin{tabular}[c]{@{}l@{}}Public\\ Dataset\end{tabular}} & \multicolumn{1}{r}{\begin{tabular}[c]{@{}l@{}}Average\\ Accuracy\end{tabular}} & \multicolumn{1}{r}{\begin{tabular}[c]{@{}l@{}}Average\\ Accuracy\end{tabular}} \\ \midrule
MNIST & FASHION & 96.69\% & \textbf{97.76\%} \\
FASHION & MNIST & \textbf{85.83\%} & 84.42\% \\
KMNIST & FASHION & 84.02\% & \textbf{86.43\%} \\
CIFAR-10 & CIFAR-100 & 67.34\% & \textbf{78.02\%} \\
CIFAR-10 & SVHN & 20.38\% &  \\ \bottomrule
\end{tabular}
\caption{Performance of FedZKT and FedMD under IID on-device data distribution.}
\label{tab:iid}
\end{table}

\begin{figure}[!tb]
\centering
\includegraphics[width=0.7\linewidth]{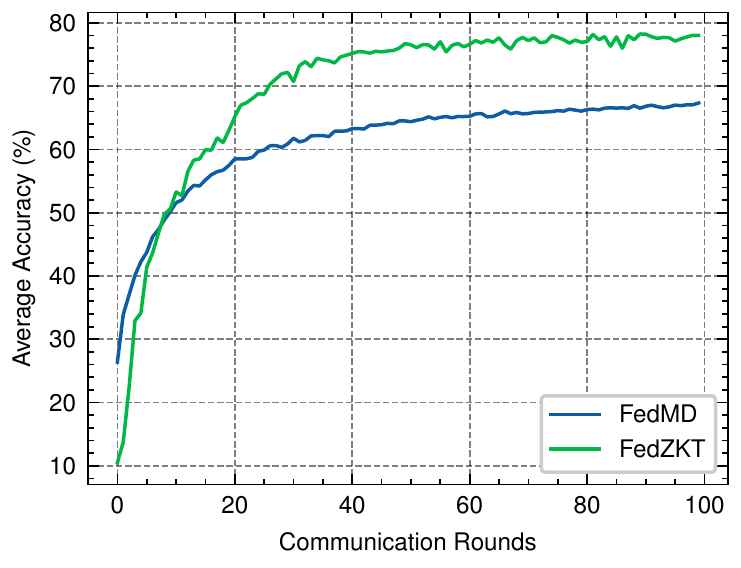}
\caption{{Learning curves of FedZKT and FedMD} (CIFAR-10, IID).}\label{fig:learning_curve}
\end{figure}

We first report the average accuracy of the global model eventually learned by the proposed FedZKT and the baseline approach FedMD under the iid on-device data distribution. As illustrated in Table~\ref{tab:iid}, FedZKT achieves higher accuracy than FedMD in most cases, which indicates the effectiveness of FedZKT. Besides, we observe that the performance of FedMD depends on the selection of the public dataset. Specifically, FedMD uses two different public datasets, CIFAR-100 and SVHN, to train on the CIFAR-10 dataset, respectively. When the public dataset (CIFAR-100) has similar distribution as the on-device dataset (CIFAR-10), FedMD achieves a higher accuracy; while when the public dataset (SVHN) is quite different from the on-device dataset (CIFAR-10), the performance of FedMD drops significantly. Thus, it is critical for FedMD to select a proper public dataset at the server, which unfortunately is extremely challenging in practice since the server may have no access to the private on-device dataset. Instead, FedZKT provides a data-free approach with even higher accuracy performance.

\begin{figure*}[!tb]
\centering
\begin{subfigure}{0.24\linewidth}
\centering
\includegraphics[width=\linewidth]{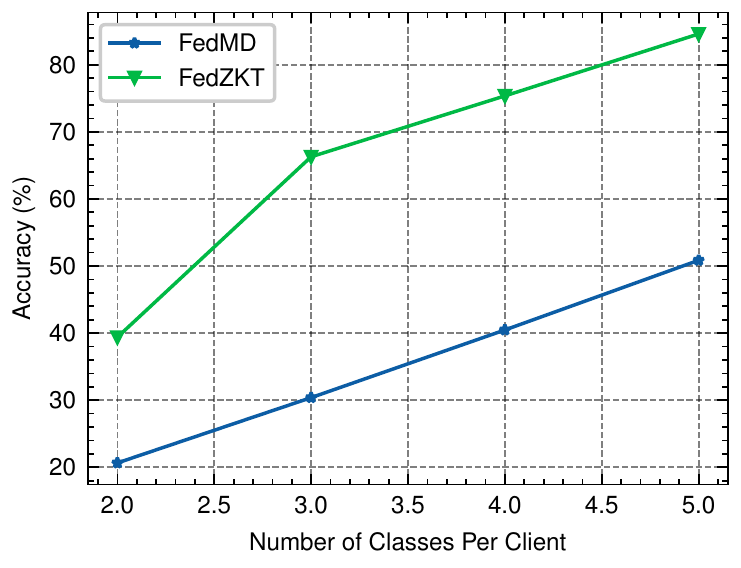}
\caption{MNIST}
\end{subfigure}
\begin{subfigure}{0.24\linewidth}
\centering
\includegraphics[width=\linewidth]{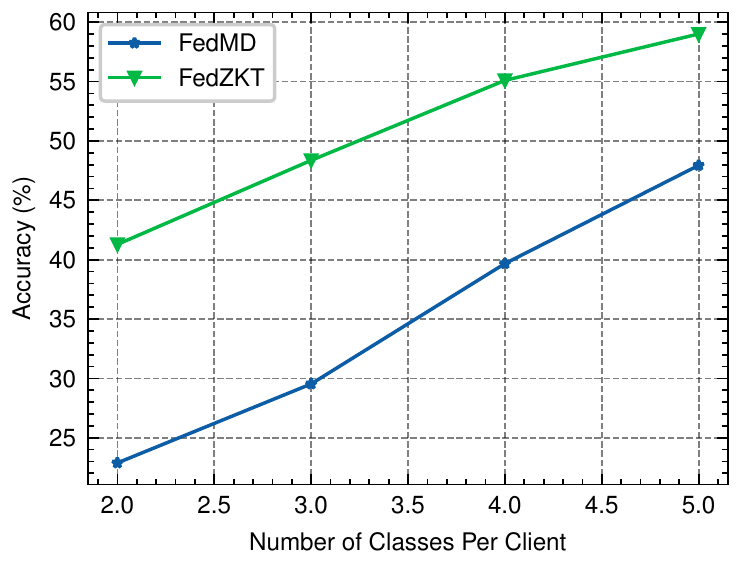}
\caption{FASHION}
\end{subfigure}
\begin{subfigure}{0.24\linewidth}
\centering
\includegraphics[width=\linewidth]{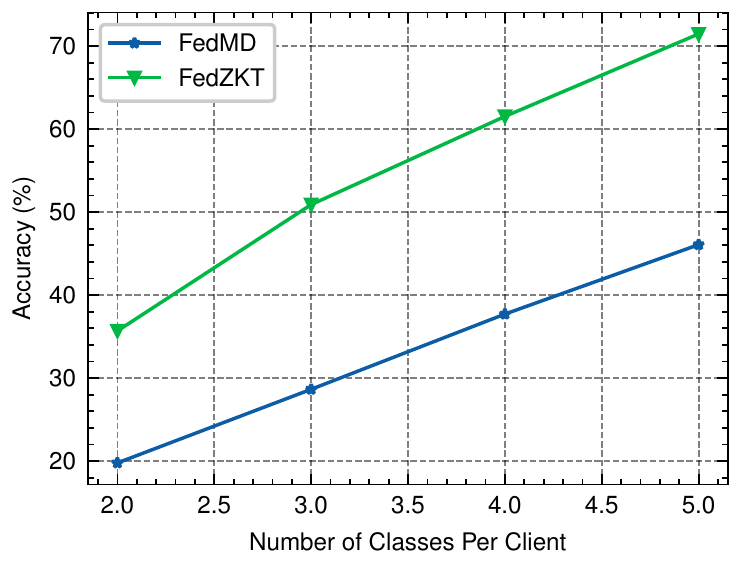}
\caption{KMNIST}
\end{subfigure}
\begin{subfigure}{0.24\linewidth}
\centering
\includegraphics[width=\linewidth]{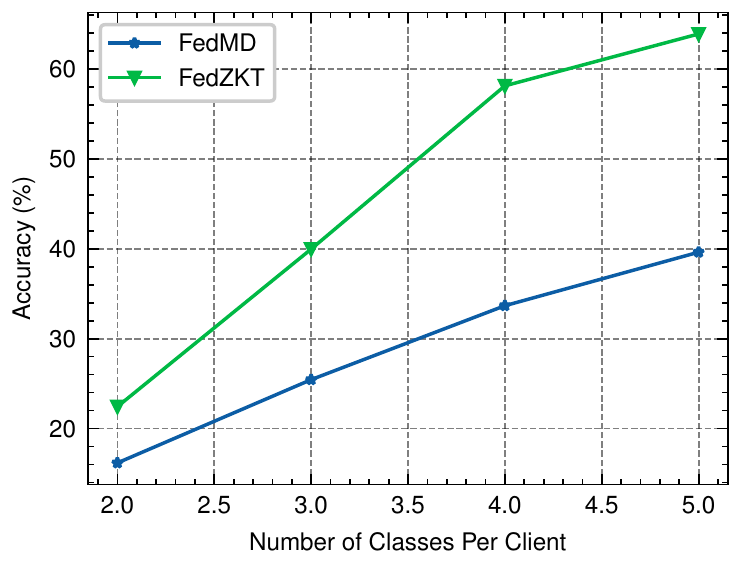}
\caption{CIFAR-10}
\end{subfigure}
\begin{subfigure}{0.24\linewidth}
\centering
\includegraphics[width=\linewidth]{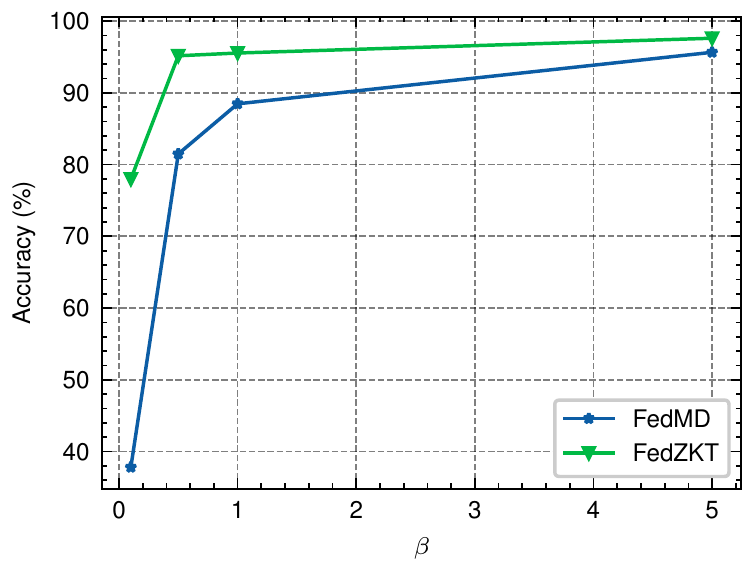}
\caption{MNIST}
\end{subfigure}
\begin{subfigure}{0.24\linewidth}
\centering
\includegraphics[width=\linewidth]{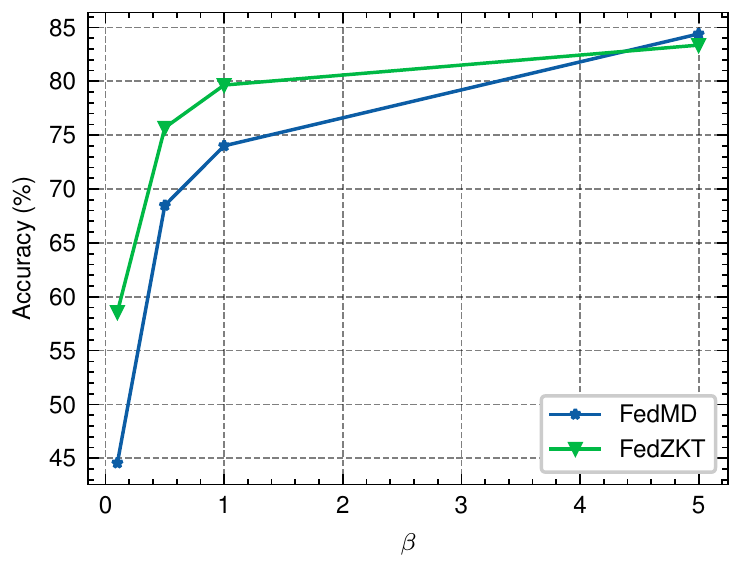}
\caption{FASHION}
\end{subfigure}
\begin{subfigure}{0.24\linewidth}
\centering
\includegraphics[width=\linewidth]{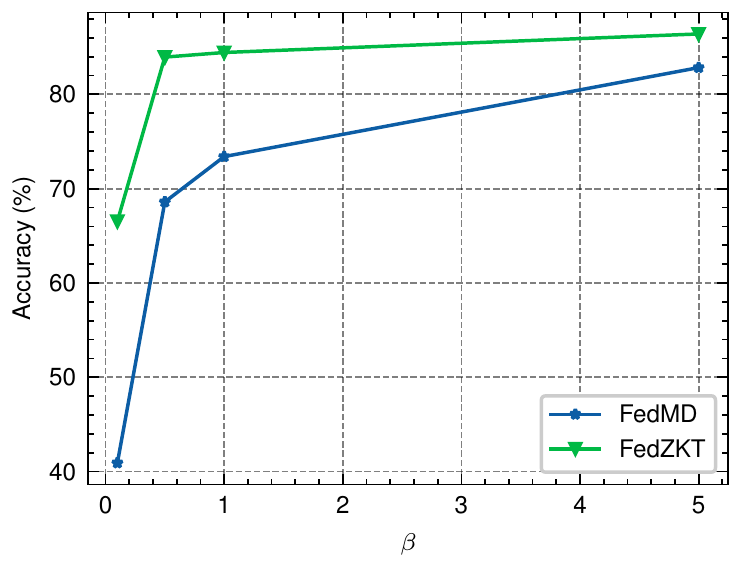}
\caption{KMNIST}
\end{subfigure}
\begin{subfigure}{0.24\linewidth}
\centering
\includegraphics[width=\linewidth]{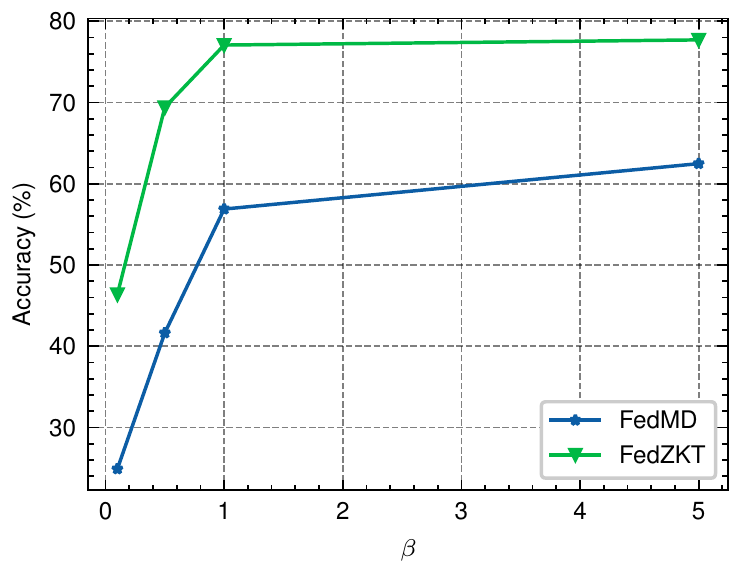}
\caption{CIFAR-10}
\end{subfigure}
\centering\caption{{Performance of FedZKT and FedMD under non-IID on-device data distribution}: Quantity-based label imbalance (a)-(d), Distribution-based label imbalance (e)-(h).}
\label{fig:quantity_noniid}
\end{figure*}

We then evaluate learning curves of FedZKT and FedMD on CIFAR-10 dataset under the iid on-device data distribution, where FedMD uses CIFAR-100 dataset as the public dataset.  As shown in Figure~\ref{fig:learning_curve}, FedMD performs better than FedZKT at the beginning of learning. Since the distribution of the public dataset (CIFAR-100) is close to that of the on-device dataset (CIFAR-10), FedMD can quickly absorb knowledge by using the public dataset at the beginning. However, we observe that with the increase of learning rounds, FedZKT eventually outperforms FedMD. This is because FedZKT can iteratively learn from on-device models to improve the generator and thus produces more representative samples, while FedMD can only use existing samples from the public dataset that cannot be improved during learning, which further indicates the benefits of our data-free design.


\subsubsection{Non-IID Data Distribution}
The accuracy performance of FedZKT and FedMD is also evaluated under the non-iid on-device data distribution. As aforementioned in Section~\ref{sec:exp:datahetero}, two data skew scenarios are considered for the non-iid setting: in the quantity-based label imbalance scenario, the values of the number of classes per device are set to be $c \in \{2,3,4,5\}$; in the distribution-based label imbalance scenario, four values of the concentration parameter $\beta \in \{0.1, 0.5, 1, 5\}$ are considered. Figure~\ref{fig:quantity_noniid} illustrates the accuracy performance under the above two non-iid data scenarios. We observe that FedZKT outperforms FedMD in almost all non-iid environments, which indicates the robustness of FedZKT in handling non-iid on-device data scenarios. 

\subsection{Ablation Studies}
This section performs ablation studies to evaluate the impact of the key modules/considerations for FedZKT under challenging federated learning environments.

\begin{table}[!tb]
\centering
\begin{tabular}{@{}llll@{}}
\toprule
Non-IID scenario & KL-divergence & $\ell_1$ norm & SL loss \\ \midrule
$C=5$ & 48.23\% & 14.60\% & \textbf{63.89\%} \\
$\beta=0.5$ & 66.17\% & 16.34\% & \textbf{69.39\%} \\ \bottomrule
\end{tabular}
\caption{Effect of loss functions for zero-shot knowledge distillation in FedZKT (CIFAR-10, Non-IID).}
\label{tab:loss}
\end{table}

\vspace{0.2em}
\subsubsection{Effects of Loss Function Design}\label{sec:loss}
We first evaluate the loss function design for zero-shot federated knowledge distillation in Section~\ref{sec:loss_kl}. We consider the more challenging non-iid on-device data scenarios: the distribution-based label imbalance with $\beta=0.5$ and the quantity-based label imbalance with $c=5$. As illustrated in Table~\ref{tab:loss}, the proposed SL loss achieves better accuracy performance than the KL-divergence loss and the $\ell_1$ norm loss in the two non-iid scenarios. In addition, our results show that $\ell_1$ norm loss is not suitable for zero-shot federated distillation under non-iid settings due to the unstable learning performance, although it can avoid the gradient vanishing in zero-shot distillation.

\vspace{0.1em}
\subsubsection{Effects of On-Device Model Architectures}\label{sec:exp:modelarch}
\begin{figure}[!tb]
\centering
\centering
\includegraphics[width=0.7\linewidth]{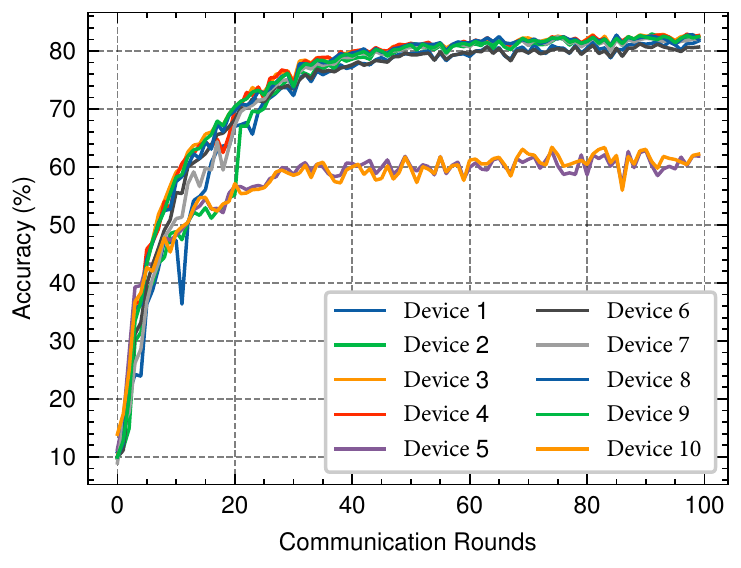}
\caption{{Effect of on-device model architecture: learning curves of devices in FedZKT with different models (CIFAR-10, IID).}}\label{fig:learning_curve_client}
\end{figure}

One key consideration of FedZKT is to enable each devices, especially the resource-constrained one, to independently design the on-device model. Hence, Figure~\ref{fig:learning_curve_client} evaluates the impact of the on-device model architecture on the learning curve of each device. Specifically, we consider ten devices under the same model architecture configuration as Table~\ref{tab:bound_model}. Detailed model settings are provided in the Appendix. Since Device 5 and Device 10 use Model E, \ie, a simple LeNet-like neural network, their on-device learning performance is lower than those using the ShuffleNetV2 and MobileNetV2 models. In addition, we present the lower bound and upper bound performance of on-device training in Table~\ref{tab:bound_model}, where the lower bound considers the on-device model is trained on its own data only; the upper bound assumes the on-device model can access others' local data. By comparing Table~\ref{tab:bound_model} with the eventual accuracy performance in Figure~\ref{fig:learning_curve_client}, we observe that the performance of FedZKT is very close to the upper-bound values, which indicates the effectiveness of FedZKT in handling federated learning with heterogeneous on-device models. 


\begin{table}[!tb]
\centering
\begin{tabular}{@{}lccccc@{}}
\toprule
Model Architecture    & Upper Bound & Lower Bound \\ \midrule
Device 1: Model A      & 84.18\%     & 50.17\%     \\
Device 2: Model B      & 86.98\%     & 54.08\%     \\
Device 3: Model C      & 88.63\%     & 58.56\%     \\
Device 4: Model D      & 87.36\%     & 57.84\%     \\
Device 5: Model E      & 70.77\%     & 58.90\%     \\
Device 6: Model A     & 84.39\%     & 50.13\%     \\
Device 7: Model B      & 85.99\%     & 51.07\%     \\
Device 8: Model C      & 88.34\%     & 62.15\%     \\
Device 9: Model D      & 88.87\%     & 59.88\%     \\
Device 10: Model E      & 70.65\%     & 54.81\%     \\ \bottomrule
\end{tabular}
\caption{Effect of on-device model architecture: lower and upper bound of on-device performance (CIFAR-10, IID).}\label{tab:bound_model}
\end{table}

\begin{figure}[!tb]
\centering
\begin{subfigure}{0.49\linewidth}
\centering
\includegraphics[width=\linewidth]{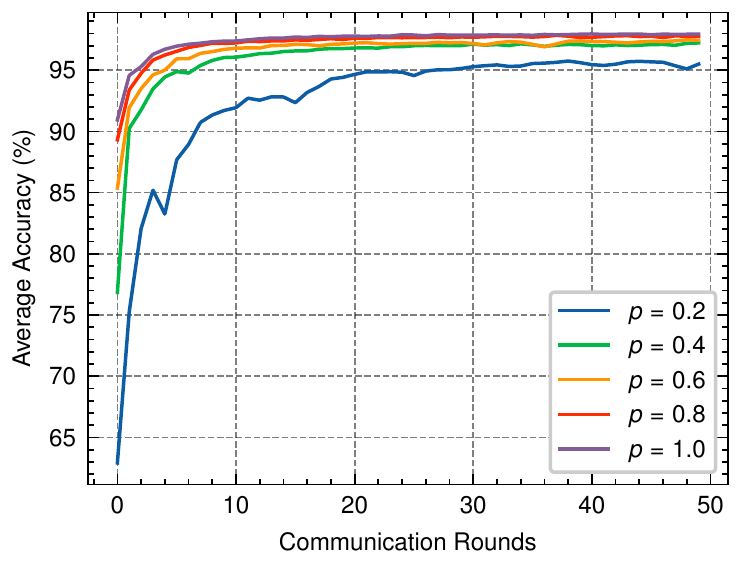}
\caption{MNIST, IID.}
\end{subfigure}
\begin{subfigure}{0.49\linewidth}
\centering
\includegraphics[width=\linewidth]{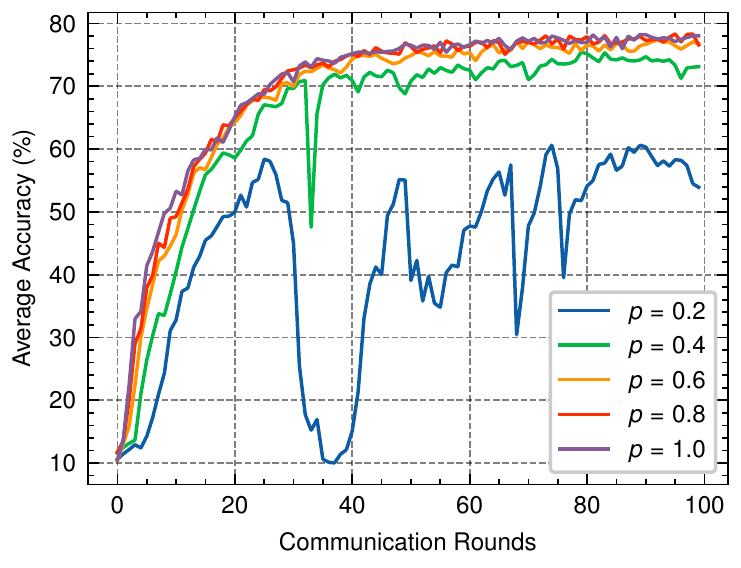}
\caption{CIFAR-10, IID.}
\end{subfigure}
\caption{{Effect of stragglers:} average accuracy of FedZKT when $p$ portion of devices are trained in each round.}
\label{fig:straggler}
\end{figure}

\vspace{0.2em}
\subsubsection{Effects of Stragglers}
Straggler effect has been one major concern in federated learning, where one or several devices cannot timely participate in training due to the unstable on-device conditions, such as the poor networking and the low battery conditions. Hence, we evaluate the effect of stragglers in FedZKT. Specifically, in each communication round, we randomly select a portion $p$ of devices as the active ones who participate in federated learning, $p\in\{0.2, 0.4, 0.6, 0.8, 1.0\}$, and the rest devices are inactive that cannot contribute to the learning round. The learning procedures follow the Algorithm~\ref{alg:fedzkt} but do not update the parameters for inactive devices. As shown in Figure~\ref{fig:straggler}, the performance of FedZKT is stable during the training in most cases. The unstable training only occurs when a very small portion of devices, \ie, $p=0.2$, participates, slowing down the process of training. As long as the majority of devices can participate in the training, the stragglers do not have a significant impact on FedZKT. 

\begin{table}[!tb]
\centering
\begin{tabular}{@{}lrr@{}}
\toprule
Non-IID scenario & no regularization & $\ell_2$ regularization \\ \midrule
$C=5$ & 56.58\%& \textbf{63.89\%}  \\
$\beta=0.5$ & 66.17\% & \textbf{69.39\%} \\ \bottomrule
\end{tabular}
\caption{Effect of $\ell_2$ regularization in FedZKT (CIFAR-10, Non-IID).}\label{tab:prox}
\end{table}

\vspace{0.2em}
\subsubsection{Effects of $\ell_2$ Regularization}\label{sec:l2_reg}
Experiments are conducted to evaluate the performance of the $\ell_2$ regularization for on-device training against the non-iid data distribution in FedZKT (Section~\ref{sec:l2reg}). As illustrated in Table~\ref{tab:prox}, the performance of using $\ell_2$ regularization is better than that without the regularization in  both quantity-based ($C$ = 5) and distribution-based ($\beta$ = 0.5) label imbalance non-iid scenarios, which demonstrates the effectiveness of the $\ell_2$ regularization.

\begin{figure}[!tb]
\centering
\begin{subfigure}{0.49\linewidth}
\centering
\includegraphics[width=\linewidth]{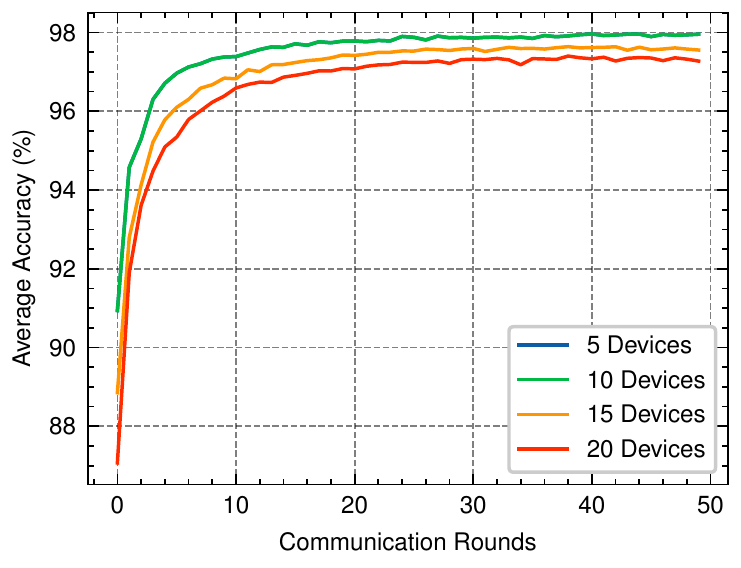}
\caption{MNIST, IID.}
\end{subfigure}
\begin{subfigure}{0.49\linewidth}
\centering
\includegraphics[width=\linewidth]{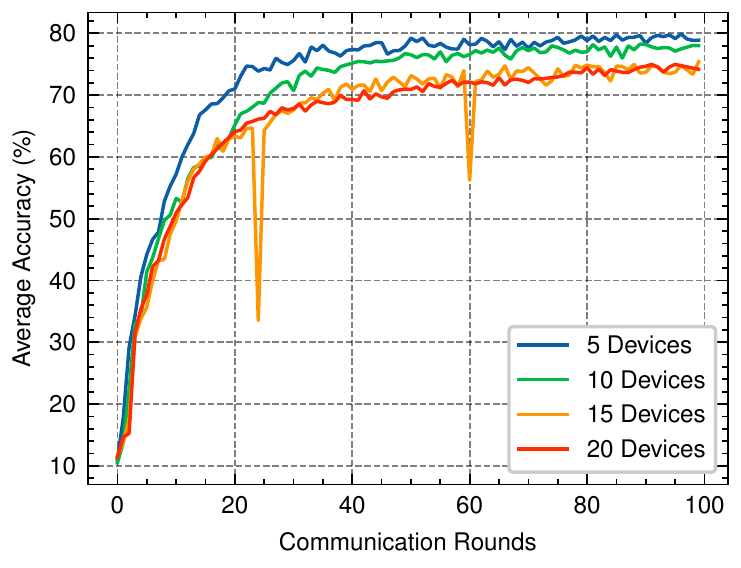}
\caption{CIFAR-10, IID.}
\end{subfigure}
\caption{{Effects of device number:} average accuracy of $K$ on-devices models participated in FedZKT ($K\in\{5, 10, 15, 20\}$). }
\label{fig:client}
\end{figure}

\vspace{0.2em}
\subsubsection{Effects of Device Number}
We finally evaluate the effect of device number $K$ on FedZKT, where $K\in\{5, 10, 15, 20\}$. Figure~\ref{fig:client} illustrates the learning curves of FedZKT for MINIST and CIFAR-10 datasets under iid conditions.  We observe that the number of devices does not affect the overall training too much. Although a smaller number of devices, \eg, $K=5$, produce higher average accuracy, FedZKT achieves good performance with more devices as well. As observed from the experiments, the effect of device numbers on FedZKT is subtle, \ie, around $\pm2\%$ in terms of the average accuracy.

\section{Conclusions}
This paper proposed an innovative federated learning framework for resource-constrained and heterogeneous devices via zero-shot knowledge transfer, named by FedZKT. FedZKT allows devices, especially the resource-constrained ones, to determine their on-device models independently. Unlike certain prior research relied on the prerequisite for private on-device knowledge, FedZKT enables knowledge transfer across heterogeneous on-device models in a data-free manner, where a new loss function, SL loss, is proposed to facilitate the zero-shot federated knowledge distillation. Moreover, to meet the unbalanced resources between the server and device sides, FedZKT assigns the compute-intensive distillation task to the server. The distilled central knowledge is then sent back in the form of on-device model parameters, which can be easily absorbed at devices. Extensive experiments demonstrate the effectiveness and the robustness of FedZKT towards on-device knowledge agnostic (data-free), on-device model heterogeneity, and other challenging federated learning scenarios, such as heterogeneous on-device data and straggler effects.



\section*{Acknowledgment}
The authors thank all the anonymous reviewers for their
insightful feedback. This work was supported by the National Science Foundation under Grant No.~2106754

\bibliographystyle{IEEEtran}
\bibliography{reference}

\section*{Appendix: Loss Function Design} 
This section justifies the two hypotheses proposed for the loss function design of zero-shot knowledge distillation in section~\ref{sec:loss_kl}. The goal of this design is to select the loss function in (\ref{equ_minmaxKD}), which measures the disagreement between the global model $\mcf$ and the ensemble of on-device models $f_{ens}$, \ie, $f_{\mathrm{ens}}(x) =\frac{1}{|\mck|}\sum_k f_k(\mbx; w_k)$, in a zero-shot approach. 

Given the input data $x$, we define the outputs of $\mcf$ after the softmax layer as $U = \mcf(x)$, and the logit outputs of the on-device model $f_k$, $k\in\mathcal{K}$ as $V_k = f_k(x)$. In addition, define the logit output before the softmax layer as $u$ and $v_k$ for $\mcf$ and $f_k$, $k\in\mathcal{K}$, respectively. Thus, we have $U=\mathrm{softmax}(u)$ and $V_k = \mathrm{softmax}(v_k)$. Based on the above definitions, in the following, we first provide the norm of gradients for $\ell_1$ norm loss, KL-divergence loss, and SL loss, respectively, followed by the hypotheses justification.

\subsection{Norm of gradients for $\ell_1$ norm loss}
Given the $\ell_1$ norm loss in our setting can be defined by
\begin{equation*}
   \mathcal{L}_{\ell_1} = ||u-\frac{1}{|\mck|}\sum_k v_k||_1,
\end{equation*}
we have the gradients of the $\ell_1$ norm loss with respect to input data $x$ by
\begin{align*}
    \nabla_x\mathcal{L}_{\ell_1} = &\sum_i\mathrm{sign}(u^i-\frac{1}{|\mck|}\sum_j v_j^i)(\frac{\partial u^i}{\partial x} - \frac{\partial(\frac{1}{|\mck|}\sum_k v_k^i)}{\partial x}),\\
    =& \sum_i\mathrm{sign}(u^i-\frac{1}{|\mck|}\sum_j v_j^i)\frac{1}{|\mck|}\sum_k(\frac{\partial u^i}{\partial x} - \frac{\partial v_k^i}{\partial x}),\\
    =&\frac{1}{|\mck|}\sum_i\sum_k \mathrm{sign}(u^i-\frac{1}{|\mck|}\sum_j v_j^i)(\frac{\partial u^i}{\partial x} - \frac{\partial v_k^i}{\partial x}),
\end{align*}
where $i$ denotes the dimension of output.
Hence, the norm of gradients of $\ell_1$ norm loss can be given by
\begin{equation}\label{eq:l1_gradient_norm}
\begin{split}
     &||\nabla_x\mathcal{L}_{\ell_1}|| = \\
     & \frac{1}{|\mck|}||\sum_i\sum_k \mathrm{sign}(u^i-\frac{1}{|\mck|}\sum v_k^i)(\frac{\partial u^i}{\partial x} - \frac{\partial v_k^i}{\partial x})||.
\end{split}
\end{equation}

\subsection{Norm of gradients for KL-divergence loss}
Given the KL-divergence loss in our setting defined by
\begin{equation*}
    \mathcal{L}_{\mathrm{KL}} = \sum_i U^i \log\frac{U^i}{\frac{1}{|\mck|}\sum_k V_k^i},
\end{equation*}
we have the gradients of KL-divergence loss with respect to input $x$, which can be given by
\begin{equation}
\begin{split}
    & \nabla_x\mathcal{L}_{\mathrm{KL}} = \\
    & \sum_i \frac{\partial U^i}{\partial x} \log \frac{U^i}{\frac{1}{|\mck|}\sum_k V_k^i} - \frac{\partial (\frac{1}{|\mck|}\sum_k V_k^i)}{\partial x} \frac{U^i}{\frac{1}{|\mck|}\sum_k V_k^i}.
\end{split}
\end{equation}

When $U$ converges to the ensemble of $V_k$, we have
\begin{equation}
\label{eq:kl_gradient}
\quad U(x) = \frac{1}{|\mck|}\sum_k V_k(x) (1+\delta(x)),
\end{equation}
where $\delta(x){\rightarrow} 0$ when $F\rightarrow f_{\mathrm{ens}}$. 
Thus, the gradients of KL-divergence loss in Eq.~\ref{eq:kl_gradient} can be given by
\begin{align}
\label{eq:kl_gradient_norm}
    \nabla_x\mathcal{L}_{\mathrm{KL}} \approx & \sum_i \frac{\partial U^i}{\partial x}\delta_i - \frac{\partial (\frac{1}{|\mck|}\sum_k V_k^i)}{\partial x}(1+\delta_i) \\
    &\quad (\mathrm{since} \log(1+\delta)\underset{\delta\rightarrow 0}{\rightarrow} \delta)\nonumber\\
    = & \sum_i \delta_i (\frac{\partial U^i}{\partial x} - \frac{\partial (\frac{1}{|\mck|}\sum_k V_k^i)}{\partial x}) \\ 
    &\quad(\forall k, \sum_i \frac{\partial V_k^i}{\partial x} = 0, \mathrm{since} \sum_i V_k^i = 1), \nonumber\\
    = & \frac{1}{|\mck|} \sum_i \sum_k \delta_i(\frac{\partial U^i}{\partial x} - \frac{\partial V_k^i}{\partial x})
\end{align}

Thus, the norm of the gradients of KL-divergence loss can be given by
\begin{equation}
    ||\nabla_x\mathcal{L}_{\mathrm{KL}}|| \approx \frac{1}{|\mck|} || \sum_i \sum_k \delta_i(\frac{\partial U^i}{\partial x} - \frac{\partial V_k^i}{\partial x})||.
\end{equation}

\subsection{Norm of gradients for SL Loss}
The proposed SL loss is defined by
\begin{equation*}
   \mathcal{L}_{\mathrm{SL}} = ||U-\frac{1}{|\mck|}\sum_k V_k||_1.
\end{equation*}

Similarly to the $\ell_1$ norm loss, the gradients of SL loss with respect to input $x$ can be given by
\begin{equation}
\begin{split}
    &\nabla_x\mathcal{L}_{\mathrm{SL}} = \\
    &\frac{1}{|\mck|}\sum_i\sum_k \mathrm{sign}(U^i-\frac{1}{|\mck|}\sum_j V_j^i)(\frac{\partial U^i}{\partial x} - \frac{\partial V_k^i}{\partial x}).
\end{split}
\end{equation}

Thus, the norm of gradients of the SL loss can be given by
\begin{equation}\label{eq:sl_gradient_norm}
\begin{split}
    &||\nabla_x\mathcal{L}_{\mathrm{SL}}|| = \\
    & \frac{1}{|\mck|}||\sum_i\sum_k \mathrm{sign}(U^i-\frac{1}{|\mck|}\sum_j V_j^i)(\frac{\partial U^i}{\partial x} - \frac{\partial V_k^i}{\partial x})||.
\end{split}
\end{equation}

\subsection{Justification of Two Hypotheses}
Based on the above norm of gradients, this section justifies the Hypothesis 1 and Hypothesis 2. We first introduce the two Lemmas provided in~\cite{truong2021data}. 

\begin{lemma}\label{lm:jacobian}
{[Lemma 2 in~\cite{truong2021data}]}
If $S(x)\in(0,1)^K$ is the softmax output of a differentiable function (\eg, a neural network) on an input $x$, $s$ is the corresponding logits vector, and $J$ is the Jacobian matrix$\frac{\partial S}{\partial s}$, then for any vector $z$, we have:
\begin{equation}
\forall z, \quad ||Jz||\leq||z||.
\end{equation}
\end{lemma}

\begin{lemma}\label{lm:convergence} 
{[Lemma 3 in~\cite{truong2021data}]} Let $U(x)$ and $V(x)$ be the softmax output of two differential functions (\eg, neural networks) on input $x$, with respective logits $u(x)$ and $v(x)$. When $U$ converges to $V$ then $\frac{\partial U}{\partial u}$ converges to $\frac{\partial V}{\partial v}$. 
\end{lemma}

Based on the above two Lemmas, we then reformulate the derived norm of gradients of the three loss functions, and finally derive the following two hypotheses. Their proofs are given below. 

\begin{table*}[!tb]
\begin{tabular}{|c|c|c|c|c|c|c|c|c|}
\hline
\multicolumn{2}{|c|}{Model A} & \multicolumn{2}{|c|}{Model B}  & \multicolumn{2}{|c|}{Model C} & \multicolumn{2}{|c|}{Model D}  & Model E \\ \hline
Arch & Parameter & Arch & Parameter & Arch & Parameter & Arch & Parameter & Arch \\ \hline
ShuffleNetV2 & net size 0.5 & ShuffleNetV2 & net size 1.0 & MobileNetV2 & width multiplier 0.8 & MobileNetV2 & width multiplier 0.6 & LeNet \\ \hline
\end{tabular}
\caption{Model Architecture for the CIFAR-10 dataset.}
\label{tab:cifar_arch}
\end{table*}

\begin{hypothesis}
When the global model $F$ converges to the ensemble of on-device models $f_{\mathrm{ens}}$, the gradients of KL divergence loss with respect to the input data $\mbx$ are smaller than those of the SL loss:
\begin{equation}
||\nabla_{\mbx}\mathcal{L}_{\mathrm{KL}}(\mbx)|| \underset{F\rightarrow f_{\mathrm{ens}}}{\leq}  ||\nabla_{\mbx}\mathcal{L}_{\mathrm{SL}}(\mbx)||.
\end{equation}
\end{hypothesis}

\begin{hypothesis}
When the global model $F$ converges to the ensemble of on-device models $f_{\mathrm{ens}}$, the gradients of the $\ell_1$ norm loss with respect to the input data $\mbx$ are greater than those of the SL loss:
\begin{equation}
||\nabla_\mbx\mathcal{L}_{\ell_1}(\mbx)|| \underset{F\rightarrow f_{\mathrm{ens}}}{\geq}  ||\nabla_{\mbx}\mathcal{L}_{\mathrm{SL}}(\mbx)||.
\end{equation}
\end{hypothesis}

\begin{proof}
Each term in the norm of gradients of $\ell_1$ norm loss in  Eq.~\ref{eq:l1_gradient_norm} can be given by
\begin{equation}
\label{eq:l1_ineq}
\forall i, k \quad ||\mathrm{sign}(u^i-\frac{1}{|\mck|}\sum_j v_j^i)(\frac{\partial u^i}{\partial x} - \frac{\partial v_k^i}{\partial x})|| = ||\frac{\partial u^i}{\partial x} - \frac{\partial v_k^i}{\partial x}||.
\end{equation}

For the SL loss, when the global model converges to the ensemble of the on-device models, each term in the norm of gradients in Eq.~\ref{eq:sl_gradient_norm} can be given by
\begin{align}
\forall i, k \quad &||\mathrm{sign}(U^i-\frac{1}{|\mck|}\sum_j V_j^i)(\frac{\partial U^i}{\partial x} - \frac{\partial V_k^i}{\partial x})|| \\ 
= & \quad ||\frac{\partial U^i}{\partial x} - \frac{\partial V_k^i}{\partial x}|| \label{eq:sl_ineq1}\\
= & \quad ||\frac{\partial U^i}{\partial u}\frac{\partial u^i}{\partial x} - \frac{\partial V_k^i}{\partial v}\frac{\partial v_k^i}{\partial x}||,\nonumber\\
\approx &  \quad||\frac{\partial U^i}{\partial u} (\frac{\partial u^i}{\partial x} - \frac{\partial v_k^i}{\partial x})|| \quad (\mathrm{Lemma}~\ref{lm:convergence}),\nonumber\\
\leq &  \quad||\frac{\partial u^i}{\partial x} - \frac{\partial v_k^i}{\partial x}|| \quad (\mathrm{Lemma}~\ref{lm:jacobian}),
\end{align}
where the last two derivations follow the Lemma 2 and Lemma 1, respectively.
Thus, by summing up the terms over $i$ and $k$, we can expect the gradients of the SL loss is smaller than the $\ell_1$ norm loss. This completes the justification for Hypothesis~\ref{hp:l1}. However, to rigorously proof this, we need further assumptions on the data distribution and models.

For the KL-divergence loss, when the global model converges to the ensemble of the on-device models, each term in the norm of gradients in Eq.~\ref{eq:kl_gradient_norm} can be given by
\begin{equation}
\label{eq:kl_eq}
\forall i, k \quad ||\delta_i (\frac{\partial U^i}{\partial x} - \frac{\partial V_k^i}{\partial x})||.
\end{equation}

Thus, by comparing Eq. \ref{eq:sl_ineq1} and Eq. \ref{eq:kl_eq}, we can expect the gradients of KL-divergence loss is smaller than that of the SL loss This completes the justification for Hypothesis~\ref{hp:kl}.
\end{proof}

\section*{Appendix: On-Device Model Architectures}
To make the experimental validation concise and easy to follow, we detail the setting of on-device model architectures in Section~\ref{sec:exp:modelarch} here. 

For the ten devices trained with the CIFAR-10 dataset, we use ShuffleNetV2 and MobileNetV2 with different numbers of filters and a LeNet-like model architectures to meet diverse device capacity for on-device models\footnote{We implement the models based on \url{https://github.com/kuangliu/pytorch-cifar}.}. 
Table~\ref{tab:cifar_arch} presents the detailed model architectures. We use SGD with a learning rate $0.01$ and a weight decay of $0.0005$. The batch size and the total training epoch are set to be $256$ and $100$, respectively.

\end{document}